\documentclass[a4paper,fleqn]{cas-dc}

\usepackage[numbers,sort&compress]{natbib}
\usepackage{type1ec}
\usepackage{amsmath,amssymb,amsfonts}
\usepackage{booktabs}
\usepackage{textcomp}
\usepackage{xcolor}

\begin{document}

\let\WriteBookmarks\relax
\def\floatpagepagefraction{1}
\def\textpagefraction{.001}

\shorttitle{GAD-MambaUNet for Lightweight Medical Image Segmentation}
\shortauthors{Wang et~al.}

\title[mode=title]{GAD-MambaUNet: Direction-Group Mamba with Gradient-Adaptive DINOv3 Distillation for Lightweight Medical Image Segmentation}

\author[1]{Fang Wang}
\fnmark[1]

\author[1]{Huitao Li}
\fnmark[1]

\author[2]{Wenhan Chao}

\author[1]{Zheng Zhuo}[orcid=0000-0001-8028-5285]
\ead{zz@bipt.edu.cn}
\cormark[1]

\author[1]{Xinxin Yang}

\affiliation[1]{organization={College of Artificial Intelligence},
  addressline={Beijing Institute of Petrochemical Technology},
  city={Beijing},
  postcode={102617},
  country={People's Republic of China}}

\affiliation[2]{organization={School of Computer Science and Engineering},
  addressline={Beihang University},
  city={Beijing},
  postcode={100191, 100083},
  country={People's Republic of China}}

\fntext[fn1]{These authors contributed equally to this work.}
\cortext[cor1]{Corresponding author.}

\begin{abstract}
Lightweight medical image segmentation demands both accurate boundary delineation and efficient inference, posing a fundamental challenge for models deployed in resource-constrained clinical environments. In this paper, we propose GAD-MambaUNet, an asymmetric local--global segmentation network that combines lightweight convolutional modeling with efficient state-space contextual aggregation. Our architecture employs lightweight convolutional blocks in shallow stages to perceive local features, and introduces Direction-Group Graph Selective Scan (DG-GSS) blocks in deeper stages for efficient long-range modeling. Within each DG-GSS block, direction--group feature responses are treated as graph nodes, and structured message passing is performed prior to multi-directional fusion, enabling cross-direction and cross-subspace information interaction. To further enhance semantic representation without increasing inference cost, we incorporate a frozen pretrained DINOv3 model as a semantic teacher during training only, transferring its semantic priors to the compact student network via Gradient-Adaptive Distillation. Extensive experiments on multiple public medical image segmentation benchmarks demonstrate that GAD-MambaUNet achieves superior segmentation accuracy with favorable inference efficiency.
\end{abstract}


\maketitle

\section{Introduction}

Accurate medical image segmentation is a cornerstone of computer-aided diagnosis, treatment planning, and quantitative disease assessment. Given a medical image, the objective is to assign a pixel-level label to each region of interest, with particular emphasis on preserving fine anatomical boundaries. This task is inherently challenging: lesions and organs often appear small, irregular, poorly contrasted, or visually similar to surrounding tissues. A robust segmentation model must therefore integrate fine-grained local evidence---essential for boundary delineation---with sufficient global contextual information to resolve foreground--background ambiguities.

The U-Net architecture~\cite{ronneberger2015unet} established the dominant encoder--decoder paradigm, using a contracting path to extract hierarchical semantic features and symmetric skip connections to preserve spatial resolution for precise reconstruction. Subsequent CNN-based extensions have improved feature fusion, attention mechanisms, and multi-scale decoding~\cite{fan2020pranet,kim2021uacanet,chen2021transunet}. Despite these advances, convolutional networks remain inherently constrained by local receptive fields, which limits their ability to model long-range dependencies---a critical capability when boundary interpretation relies on global lesion shape, continuity cues, or contextual information.

Transformer-based architectures, such as TransUNet~\cite{chen2021transunet}, address this limitation by introducing self-attention for global feature aggregation. This design substantially enlarges the receptive field and improves contextual reasoning. However, global attention and large-scale pretrained backbones come at a cost: increased parameter counts, higher memory consumption, and slower inference, which hinder deployment in resource-limited clinical or edge-device settings. As medical AI shifts from centralized laboratory analysis to point-of-care and edge applications, segmentation accuracy alone no longer suffices. Model efficiency---measured by parameter count, FLOPs, and inference latency---has become equally important.

\begin{figure*}[!t]
\centering
\includegraphics[width=0.49\textwidth]{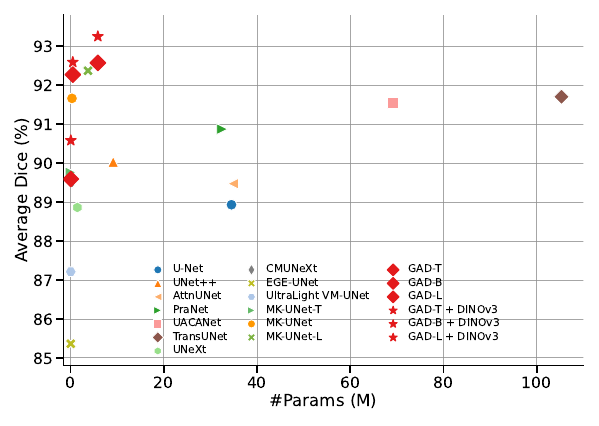}\hfill
\includegraphics[width=0.49\textwidth]{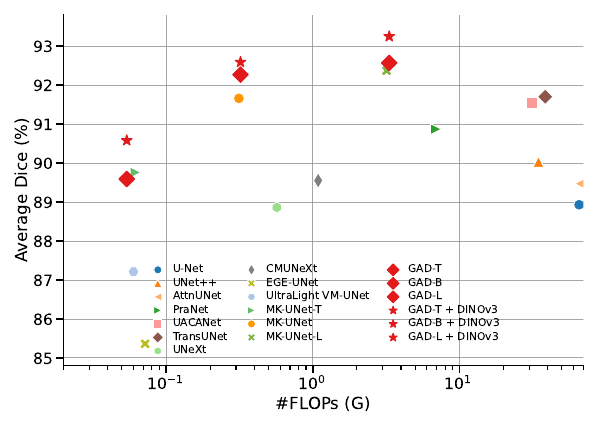}
\caption{Accuracy--efficiency comparison across PH$^2$, ISIC2018, CVC-ClinicDB, and CVC-ColonDB. The vertical axis reports the macro-average Dice score over the four datasets, while the horizontal axes show student-side parameters and FLOPs. Compared with existing lightweight segmentation networks, GAD-MambaUNet achieves a more favorable Pareto balance between segmentation accuracy and deployment cost. The DINOv3 teacher is used only during training and is excluded from deployment cost.}
\label{fig:efficiency-comparison}
\end{figure*}

This has motivated a series of lightweight segmentation networks based on depth-wise convolution~\cite{rahman2025mkunet}, compact channel designs~\cite{tang2024cmunext}, and efficient token mixing~\cite{ruan2023egeunet,rahman2024emcad,valanarasu2022unext}. Among emerging techniques, visual state-space models, particularly Mamba~\cite{gu2023mamba} and VMamba~\cite{liu2024vmamba}, offer a compelling alternative: they achieve global receptive fields with linear sequence complexity, avoiding the quadratic cost of self-attention. VM-UNet~\cite{ruan2024vmunet} further incorporates visual state-space blocks into a U-shaped segmentation backbone.

To further reduce computational cost, grouped Mamba~\cite{wu2024ultralight} partitions feature channels into parallel groups and performs multi-directional selective scanning across each group independently. While this strategy effectively lowers FLOPs, it introduces a fundamental limitation: different direction--group branches are processed in isolation and merged only after scanning. This independent design restricts the exchange of complementary contextual cues across scanning directions and channel subspaces, potentially weakening the model's capacity to capture holistic scene understanding. To address this limitation, we propose Direction-Group Graph Selective Scan (DG-GSS), which models direction--group feature responses as graph nodes and enables structured message passing before multi-directional fusion, thereby facilitating cross-direction and cross-subspace interaction.

Training lightweight segmentation models on limited medical datasets often leads to insufficient semantic abstraction. While larger vision foundation models, such as DINOv3~\cite{simeoni2025dinov3}, provide rich semantic priors, they are too heavy for deployment. Feature distillation offers a practical middle ground: a frozen teacher provides dense supervision during training, and is discarded at inference. However, a fixed distillation coefficient cannot adapt to the evolving optimization state of the student, potentially under- or over-regularizing the segmentation objective. To overcome this, we adapt Gradient-Adaptive Distillation (GAD), which dynamically adjusts the distillation strength based on the gradient contribution of the aligned decoder block, ensuring balanced semantic transfer without compromising the primary task.

Based on the above analysis, we propose GAD-MambaUNet, an asymmetric local--global lightweight segmentation network for medical image segmentation. Our architecture employs lightweight convolutional blocks in shallow stages to retain fine-grained boundary details, and introduces DG-GSS blocks in deeper stages for efficient long-range features. Within each DG-GSS block, direction--group feature responses are treated as graph nodes, and structured message passing is performed prior to multi-directional fusion, enabling cross-direction and cross-subspace information interaction. We further incorporate a frozen pretrained DINOv3 model as a semantic teacher during training only, transferring its semantic priors to the compact student network via Gradient-Adaptive Distillation.

In summary, the main contributions of this work can be categorized into four aspects:

\begin{itemize}
\item We propose DG-GSS, a novel graph-enhanced selective scan mechanism that models direction--group feature responses as graph nodes and performs structured message passing before multi-directional fusion, improving contextual interaction without increasing inference cost.

\item We introduce Gradient-Adaptive Distillation, which dynamically regulates the distillation coefficient according to the gradient share of the aligned decoder block during training, and demonstrate its effectiveness in transferring DINOv3 semantic priors to a compact student without incurring inference overhead.

\item We design GAD-MambaUNet, an asymmetric local--global lightweight segmentation network that combines convolutional local modeling in shallow stages with DG-GSS-based long-range modeling in deep stages, achieving a favorable accuracy--efficiency balance.



\item We conduct extensive experiments on PH$^2$, ISIC2018, CVC-ClinicDB, and CVC-ColonDB, showing that GAD-MambaUNet achieves a superior accuracy--efficiency balance in terms of Dice score, parameters, and FLOPs. As shown in Fig.~\ref{fig:efficiency-comparison}, the proposed models obtain more favorable performance, while DINOv3 supervision improves accuracy without increasing inference cost.

\end{itemize}

\section{Related Work}

\subsection{Medical Segmentation and Lightweight U-Shaped Networks}

U-Net~\cite{ronneberger2015unet} and its encoder--decoder variants remain a primary template for medical image segmentation because skip connections combine deep semantic features with spatially precise reconstruction. This design is particularly suitable for medical segmentation, where the model must simultaneously capture high-level anatomical or lesion semantics and recover fine boundary details. Task-specific methods strengthen this template through mechanisms such as reverse attention in PraNet~\cite{fan2020pranet} and uncertainty-aware context attention in UACANet~\cite{kim2021uacanet}. TransUNet~\cite{chen2021transunet} instead introduces a Transformer encoder to enlarge the receptive field through global self-attention. These approaches establish the value of contextual reasoning for resolving ambiguous foreground--background regions, but their attention modules or pretrained encoders can be costly when the deployment budget is limited.

Lightweight U-shaped networks reduce this cost through compact channel schedules and efficient local operators. UNeXt~\cite{valanarasu2022unext} combines convolution with shifted MLP blocks, CMUNeXt~\cite{tang2024cmunext} uses large kernels and skip fusion, EGE-UNet~\cite{ruan2023egeunet} develops group-enhanced attention, and EMCAD~\cite{rahman2024emcad} employs efficient multi-scale convolutional decoding. MK-UNet~\cite{rahman2025mkunet} further uses parallel multi-kernel depth-wise convolutions, channel shuffle, and lightweight decoder attention to capture multi-scale boundary evidence with compact cost. These lightweight CNN-based designs are effective for preserving local details and improving deployment efficiency. However, they still mainly rely on local or convolution-dominated operators and may lack explicit long-range contextual modeling when target boundaries are incomplete, lesion regions are ambiguous, or distant anatomical structures share similar textures. This limitation motivates the use of more efficient global modeling mechanisms in lightweight segmentation networks. 

\subsection{Visual State-Space Models}

State-space models offer an efficient alternative to self-attention for long-range feature modeling. Mamba~\cite{gu2023mamba} uses input-dependent state transitions with linear sequence complexity, while VMamba~\cite{liu2024vmamba} extends selective scanning to visual feature maps through multiple two-dimensional traversal directions. By replacing quadratic global attention with selective scanning, visual state-space models provide a favorable trade-off between global context modeling and computational efficiency. VM-UNet~\cite{ruan2024vmunet} incorporates visual state-space blocks into a U-shaped segmentation architecture, showing the potential of state-space modeling for medical image segmentation.

For lightweight medical segmentation, UltraLight VM-UNet~\cite{wu2024ultralight} further reduces Vision Mamba cost by partitioning deep features into parallel channel groups and processing them with parallel Vision Mamba branches. This grouped design is attractive because global modeling is introduced only in a compact and computationally efficient form. However, in grouped multi-directional scans, direction--group branches are typically processed independently and merged only after selective scanning. Although this design reduces computation, it does not explicitly model the structural relationship between traversal directions and channel subspaces.

As a result, complementary directional cues and group-wise semantic responses cannot be exchanged before fusion. Different traversal directions may capture different aspects of spatial context, while different channel groups may encode distinct but correlated semantic patterns. Processing them in isolation can therefore weaken the model's ability to form a holistic contextual representation. This limitation motivates our Direction-Group Graph Selective Scan (DG-GSS), which introduces structured interaction among direction--group responses while preserving the efficiency of grouped selective scanning.

\subsection{Foundation-Model Distillation for Lightweight Segmentation}

Strong Transformer encoders and vision foundation models provide rich semantic representations for medical image segmentation. Swin Transformer~\cite{liu2021swin} and its medical segmentation variant Swin-Unet~\cite{cao2022swinunet} model hierarchical visual features, while DINOv3~\cite{simeoni2025dinov3} provides dense features through large-scale self-supervised pretraining. Such models can provide strong high-level semantic priors, which are useful for distinguishing ambiguous lesion regions from visually similar background tissues. However, directly deploying these large encoders conflicts with the parameter, memory, and latency budgets of lightweight systems.

Feature-level distillation offers a practical alternative: a frozen teacher provides dense supervision only during training, allowing a compact student to inherit semantic context without the teacher's deployment cost~\cite{hinton2015distilling}. This strategy is especially suitable for lightweight segmentation, where the student must remain efficient during inference but can still benefit from stronger semantic guidance during optimization. Most distillation settings use a fixed distillation coefficient. However, a fixed coefficient cannot adapt to the dynamic state of optimization: as training progresses, the gradient contribution of the teacher-supervised deep student feature to the overall network can change, so the same coefficient may yield insufficient teacher supervision or impose an overly strong constraint on the primary task objective.

To address this dynamic balancing problem, RT-DETRv4~\cite{liao2025rtdetrv4} proposes a gradient-aware distillation strategy for real-time detection. In its reported configuration, a frozen DINOv3-ViT-B teacher supervises deep detector features through a Deep Semantic Injector, and Gradient-guided Adaptive Modulation adjusts the semantic-transfer strength using gradient norm ratios. Inspired by this strategy, our work adapts DINOv3-based gradient-adaptive distillation to lightweight medical image segmentation. Specifically, teacher features are aligned with a deep decoder representation of the student, and the distillation strength is dynamically regulated according to the gradient share of the aligned decoder block. During inference, the teacher and projection head are discarded, so the semantic supervision improves the compact student without increasing deployment cost.

\begin{figure*}[t]
    \centering
    \includegraphics[width=\textwidth]{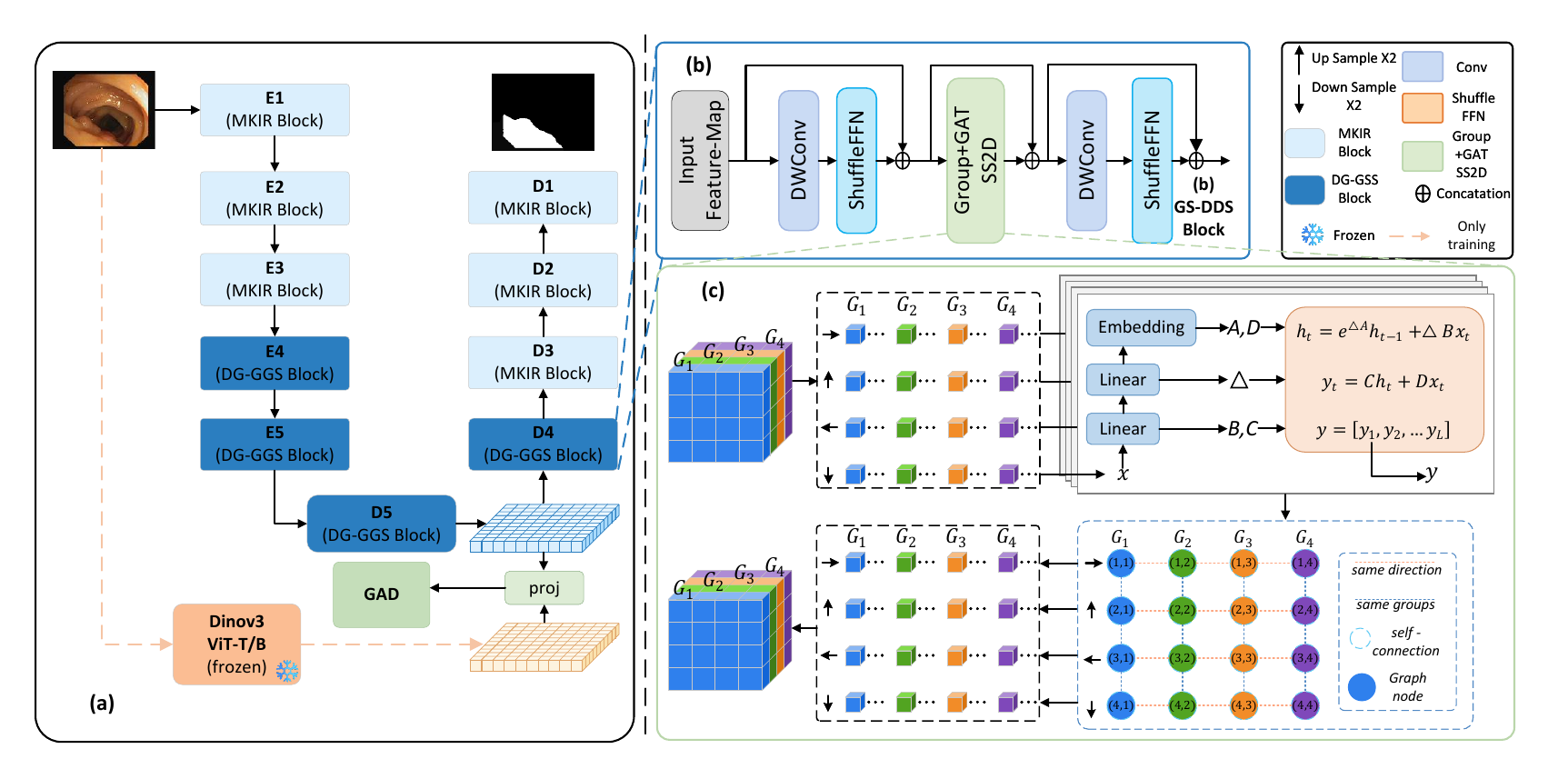}
    \caption{Overall architecture of GAD-MambaUNet. High-resolution stages retain multi-kernel local modeling, whereas deep stages use grouped state-space blocks with DG-GSS. During training, a frozen DINOv3 teacher supervises the first decoder block and GAD regulates the distillation coefficient according to its gradient share. The teacher and projection head are removed at inference.}
    \label{fig:overall-network}
\end{figure*}

\section{Methodology}
\label{sec:method}

We describe the proposed GAD-MambaUNet by first introducing its core building block and then integrating it into an asymmetric U-shaped segmentation architecture. 
As shown in Fig.~\ref{fig:overall-network}, the left part presents the overall student network and the training-time DINOv3 teacher, while the right part details the proposed block-level design and the internal structure of Direction-Group Graph Selective Scan (DG-GSS). 
We first present DG-GSS, which enables structured interaction among scan directions and channel groups. Then, we describe how DG-GSS is embedded into the overall GAD-MambaUNet architecture. Finally, we introduce the training-time DINOv3 supervision with Gradient-Adaptive Distillation (GAD), where the teacher model is used only during training and removed during inference.

\subsection{Direction-Group Graph Selective Scan}

As illustrated in the lower-right part of Fig.~\ref{fig:overall-network}, DG-GSS is designed to enhance grouped multi-directional selective scanning by introducing graph-based interaction among direction--group responses. DG-GSS regards each scan-direction and channel-group response as a graph node and performs structured message passing before the final multi-directional fusion.

Let $\mathbf{X}\in\mathbb{R}^{B\times D_i\times h\times w}$ denote the inner state-space feature of DG-GSS, where $D_i$ is the inner channel dimension. We adapt four Cross2D traversal directions and partition the inner channels into $M$ groups. With $d=D_i/M$ and $L=hw$, cross scanning produces
\begin{equation}
\mathbf{X}_{\mathrm{scan}}\in\mathbb{R}^{B\times K\times M\times d\times L},\qquad K=4.
\end{equation}

For each direction--group pair, the selective-scan parameters $(\Delta,\mathbf{B},\mathbf{C})$ are generated by independent grouped projections, and selective scanning is applied along the corresponding sequence. The directional outputs are then realigned to the original spatial coordinate system:
\begin{equation}
\mathbf{Y}\in\mathbb{R}^{B\times K\times M\times d\times h\times w}.
\end{equation}

Each tensor $\mathbf{Y}_{k,m}$ corresponds to the response of scan direction $k$ and channel group $m$. We regard every direction--group response as a graph node and obtain its node descriptor by global average pooling:
\begin{equation}
\mathbf{r}_{k,m}
=
\frac{1}{hw}
\sum_{u=1}^{h}
\sum_{v=1}^{w}
\mathbf{Y}_{k,m,:,u,v}.
\end{equation}
After layer normalization and linear projection, the node descriptor is transformed into $\mathbf{h}_{k,m}$.

In our implementation, we adopt a factorized direction-group graph, where two nodes are connected if they share the same scan direction or the same channel group, and self-connections are included. For two connected nodes $i$ and $j$, additive graph-attention logits~\cite{velickovic2018gat} are computed as
\begin{equation}
e_{ij}
=
\operatorname{LeakyReLU}
\left(
\mathbf{a}_{s}^{\top}\mathbf{h}_{i}
+
\mathbf{a}_{t}^{\top}\mathbf{h}_{j}
\right).
\end{equation}
The attention coefficients are obtained by masked softmax over the neighborhood:
\begin{equation}
\alpha_{ij}
=
\frac{\exp(e_{ij})}
{\sum_{j'\in\mathcal{N}(i)}\exp(e_{ij'})},
\qquad j\in\mathcal{N}(i).
\end{equation}

Graph-based message passing is then performed on the full spatial feature maps:
\begin{equation}
\widetilde{\mathbf{Y}}_i
=
\mathbf{Y}_i
+
\gamma
\sum_{j\in\mathcal{N}(i)}
\alpha_{ij}V(\mathbf{Y}_j),
\end{equation}
where $V(\cdot)$ is a shared $1\times1$ value projection and $\gamma$ is a learnable scaling parameter initialized to zero. This initialization makes DG-GSS start from the original grouped selective-scan behavior and progressively learn direction--group information exchange during training.

Finally, the enhanced group responses are concatenated along the channel dimension for each direction, and the four aligned directional feature maps are summed:
\begin{equation}
\mathbf{Z}
=
\sum_{k=1}^{K}
\operatorname{Concat}_{m=1}^{M}
\left(
\widetilde{\mathbf{Y}}_{k,m}
\right).
\end{equation}
Output normalization is then applied to obtain the DG-GSS output.
In this way, DG-GSS preserves the efficiency of grouped multi-directional selective scanning while explicitly modeling the structural relationship between scan directions and channel groups.

\subsection{GAD-MambaUNet Architecture}

After defining DG-GSS, we embed it into a lightweight block and integrate the block into an asymmetric U-shaped segmentation network. As shown in the upper-right part of Fig.~\ref{fig:overall-network}, the DG-GSS block consists of three residual components: a pre-scan local mixing stage, the proposed DG-GSS layer, and a post-scan local refinement stage. Given an input feature $\mathbf{x}$, the block is formulated as
\begin{equation}
\begin{aligned}
\mathbf{u}_1 &= \mathbf{x}+\operatorname{FFN}_0(\operatorname{DW}_0(\mathbf{x})), \\
\mathbf{u}_2 &= \mathbf{u}_1+\operatorname{DG\text{-}GSS}(\mathbf{u}_1), \\
\mathbf{y} &= \mathbf{u}_2+\operatorname{FFN}_1(\operatorname{DW}_1(\mathbf{u}_2)),
\end{aligned}
\end{equation}
where $\operatorname{DW}(\cdot)$ denotes depth-wise convolution for local mixing and $\operatorname{FFN}(\cdot)$ denotes a lightweight feed-forward refinement module. The local operators enhance short-range texture and channel interactions, while DG-GSS provides efficient long-range context aggregation through direction--group graph interaction.

The overall GAD-MambaUNet follows a five-level encoder--decoder architecture, as shown in the left part of Fig.~\ref{fig:overall-network}. Given an input image $\mathbf{x}\in\mathbb{R}^{3\times H\times W}$, the network predicts a binary segmentation logit map $\mathbf{z}\in\mathbb{R}^{1\times H\times W}$. In the base configuration, the channel widths are set to $[16,32,64,96,160]$. To balance local boundary preservation and global context modeling, the first three encoder stages and the last three decoder stages use lightweight multi-kernel inverted residual (MKIR) blocks inherited from MK-UNet~\cite{rahman2025mkunet}, while the two deepest encoder stages and the first two decoder stages use DG-GSS blocks. Resolution and channel changes are performed outside these blocks through depth-wise downsampling with point-wise projection or point-wise projection followed by bilinear upsampling, keeping the feature width fixed inside each block.

For decoder reconstruction, skip connections transfer high-resolution encoder features to the corresponding decoder stages. Before fusion, a grouped attention gate filters the skip feature according to the decoder context, reducing irrelevant background responses. Let $\mathbf{g}$ denote the upsampled decoder feature and $\mathbf{s}$ the encoder feature at the same resolution. The gated skip feature is computed as
\begin{equation}
\boldsymbol{\alpha}
=
\sigma\left(
\psi\left(
\phi(W_g\mathbf{g}+W_s\mathbf{s})
\right)
\right),
\qquad 
\hat{\mathbf{s}}=\boldsymbol{\alpha}\odot\mathbf{s}.
\end{equation}
where $W_g$ and $W_s$ are grouped convolutions, $\phi$ is ReLU, and $\psi$ maps the joint response to a spatial gate. The filtered skip feature $\hat{\mathbf{s}}$ is then fused with the decoder feature. The first decoder block operates at the deepest decoder level and provides the aligned student feature for the training-time DINOv3 supervision described in the next subsection.

\subsection{Training-Time DINOv3 Supervision with GAD}

Although DG-GSS improves contextual modeling inside the compact student network, training a lightweight segmentation model on limited medical datasets may still lead to insufficient semantic abstraction. Inspired by the DINOv3-based gradient-adaptive distillation strategy in RT-DETRv4~\cite{liao2025rtdetrv4}, we adapt training-time foundation-model supervision to lightweight medical image segmentation. A frozen pretrained DINOv3 model~\cite{simeoni2025dinov3} is used as the semantic teacher only during training, and the teacher branch is removed during inference.

We adopt a capacity-matched teacher assignment strategy for different student scales. Specifically, DINOv3 ViT-T/16 is used for the Tiny, Small, and Base variants, while DINOv3 ViT-B/16 is used for the Medium and Large variants. For compact students, the smaller teacher provides moderate semantic guidance while avoiding an excessive teacher--student capacity gap and unnecessary training overhead. For larger students, the stronger ViT-B/16 teacher provides richer high-level semantic supervision, which can be better absorbed by higher-capacity decoder representations. Since the DINOv3 teacher is discarded during inference, this assignment affects only the training-time supervision strength and does not change the student-side deployment cost.

As described in the previous subsection, the student feature used for distillation is extracted after the first decoder block and before the first upsampling operation. This feature is semantically rich, spatially compact, and computationally inexpensive for feature alignment. Let $\mathbf{F}_{s}$ denote the projected student feature, and let $\mathbf{F}_{t}$ denote the corresponding DINOv3 teacher feature. The student feature is projected to the teacher dimension by a $1\times1$ convolution followed by batch normalization when their channel dimensions are different. 

After spatial alignment, both feature maps are flattened into $N$ spatial tokens and $\ell_2$-normalized along the channel dimension. We use a cosine feature distillation loss:
\begin{equation}
\mathcal{L}_{\mathrm{dist}}
=
\frac{1}{N}\sum_{n=1}^{N}
\left[
1-
\frac{
\mathbf{f}_{s,n}^{\top}\mathbf{f}_{t,n}
}{
\left\|\mathbf{f}_{s,n}\right\|_2
\left\|\mathbf{f}_{t,n}\right\|_2
+\epsilon
}
\right]
\end{equation}
where $\mathbf{f}_{s,n}$ and $\mathbf{f}_{t,n}$ denote the $n$-th student and teacher tokens, respectively, and $\epsilon$ is a small constant for numerical stability.

The segmentation objective follows the structure-aware weighted BCE and weighted IoU loss~\cite{fan2020pranet}:
\begin{equation}
\mathcal{L}_{\mathrm{seg}}
=
\mathcal{L}_{\mathrm{wbce}}
+
\mathcal{L}_{\mathrm{wiou}}
\end{equation}
The overall training objective is
\begin{equation}
\mathcal{L}
=
\mathcal{L}_{\mathrm{seg}}
+
\lambda_e\mathcal{L}_{\mathrm{dist}}
\end{equation}
where $\lambda_e$ is the distillation coefficient at epoch $e$.

\begin{table*}[h]
    \centering
    \caption{Quantitative comparison with representative lightweight medical image segmentation methods on PH$^2$, ISIC2018, CVC-ClinicDB, and CVC-ColonDB. Dice scores are reported in \%. Baseline results are reproduced or cited from their reported settings when available. Params and FLOPs measure student-side inference complexity. For GAD-MambaUNet with DINOv3 supervision, the teacher model is used only during training and is excluded from deployment cost.}
    \label{tab:reference_results}
    \scriptsize
    \resizebox{\textwidth}{!}{%
    \begin{tabular}{lcccccccc}
        \toprule
        Method & Params(M) & FLOPs(G) & Throughput(/s) & PH$^2$ & ISIC2018 & ClinicDB & ColonDB & Avg. \\
        \midrule
        U-Net & 34.53 & 65.53 & 92.74 & 93.68 & 86.67 & 91.43 & 83.95 & 88.93 \\
        PraNet & 32.55 & 6.93 & 49.94 & 94.16 & 88.46 & 91.71 & 89.16 & 90.87 \\
        UACANet & 69.16 & 31.51 & 33.72 & 94.37 & 88.72 & 93.29 & 89.76 & 91.53 \\
        TransUNet & 105.32 & 38.52 & 50.91 & 94.63 & 89.04 & 93.18 & 89.97 & 91.70 \\
        UNeXt & 1.47 & 0.57 & 134.33 & 93.60 & 87.78 & 90.20 & 83.84 & 88.86 \\
        CMUNeXt & 0.418 & 1.09 & \textbf{136.68} & 94.02 & 87.51 & 92.82 & 83.85 & 89.55 \\
        EGE-UNet & 0.054 & 0.072 & 78.70 & 93.69 & 86.95 & 84.76 & 76.03 & 85.36 \\
        UltraLight VM-UNet & 0.050 & 0.060 & 76.38 & 93.82 & 87.85 & 87.11 & 80.06 & 87.21 \\
        MK-UNet-T & \textbf{0.027} & 0.062 & 109.76 & 94.56 & 88.19 & 91.26 & 85.03 & 89.76 \\
        MK-UNet & 0.316 & 0.314 & 107.23 & 94.39 & 88.74 & 93.48 & 90.01 & 91.66 \\
        MK-UNet-L & 3.76 & 3.19 & 96.38 & 94.56 & 89.25 & 93.85 & 91.82 & 92.37 \\
        \midrule
GAD-MambaUNet-T w/o DINOv3 & 0.076 & 0.054 & 100.14 & 94.43 & 88.27 & 92.47 & 83.20 & 89.59 \\
GAD-MambaUNet-S w/o DINOv3 & 0.199 & 0.128 & 99.65 & 94.40 & 88.63 & 93.85 & 90.17 & 91.76 \\
GAD-MambaUNet-B w/o DINOv3 & 0.493 & 0.322 & 97.97 & 94.90 & 88.91 & 94.04 & 91.24 & 92.27 \\
GAD-MambaUNet-M w/o DINOv3 & 1.839 & 0.955 & 92.25 & 94.71 & 89.13 & 94.07 & 91.35 & 92.31 \\
GAD-MambaUNet-L w/o DINOv3 & 5.871 & 3.325 & 90.86 & 94.88 & 89.34 & 94.70 & 91.37 & 92.57 \\
        \midrule
GAD-MambaUNet-T w/ ViT-T & 0.076 & \textbf{0.054} & 100.14 & 94.61 & 88.61 & 93.09 & 86.00 & 90.58 \\
GAD-MambaUNet-S w/ ViT-T & 0.199 & 0.128 & 99.65 & 94.73 & 88.94 & 94.16 & 90.76 & 92.15 \\
GAD-MambaUNet-B w/ ViT-T & 0.493 & 0.322 & 97.97 & 95.16 & 89.06 & 94.17 & 91.96 & 92.59 \\
GAD-MambaUNet-M w/ ViT-B & 1.839 & 0.955 & 92.25 & 95.21 & 89.11 & 94.52 & \textbf{93.08} & 92.98 \\
GAD-MambaUNet-L w/ ViT-B & 5.871 & 3.325 & 90.86 & \textbf{95.37} & \textbf{89.53} & \textbf{95.08} & 93.01 & \textbf{93.25} \\
        \bottomrule
    \end{tabular}
    }
\end{table*}

Using a fixed distillation coefficient assumes that the proper teacher supervision strength remains unchanged throughout training. However, the optimization state of the student changes over time, and the gradient contribution of the teacher-supervised decoder block may become either too weak or too dominant. To balance semantic supervision and the primary segmentation objective, GAD dynamically updates $\lambda_e$ according to the gradient share of the aligned decoder block. After back-propagation and before gradient clipping, we compute
\begin{equation}
q_e
=
100\times
\frac{
\sum_{\theta_i\in\Theta_a}
\left\|\nabla_{\theta_i}\mathcal{L}\right\|_1
}{
\sum_{\theta_j\in\Theta}
\left\|\nabla_{\theta_j}\mathcal{L}\right\|_1
+\epsilon
}
\end{equation}
where $\Theta_a$ denotes the parameters of the aligned decoder block and $\Theta$ denotes all trainable student parameters.

Given a target gradient-share interval $[\rho-\delta,\rho+\delta]$, GAD adjusts the next-epoch distillation coefficient only when $q_e$ falls outside this interval. Let $q^{*}$ be the nearest target boundary, and define $p_e=q_e/100$ and $p^{*}=q^{*}/100$. The odds-ratio adjustment is computed as
\begin{equation}
r_e
=
\frac{
p^{*}(1-p_e)
}{
p_e(1-p^{*})+\epsilon
}
\end{equation}
The distillation coefficient for the next epoch is updated by
\begin{equation}
\lambda_{e+1}
=
\operatorname{clip}
\left(
\lambda_e
\operatorname{clip}(r_e,r_{\min},r_{\max}),
\lambda_{\min},
\lambda_{\max}
\right)
\end{equation}
where $r_{\min}$ and $r_{\max}$ bound the per-epoch adjustment, and $\lambda_{\min}$ and $\lambda_{\max}$ constrain the overall distillation strength.

Training begins with a segmentation-only burn-in stage, followed by a linear distillation warm-up. GAD is activated after warm-up to adaptively regulate the teacher contribution. During inference, the DINOv3 teacher, the feature projection head, and the GAD update are all discarded. Therefore, the proposed training-time supervision improves the compact student without increasing deployment cost.

\section{Experiments and Results}
\label{sec:experiments}

\begin{table*}[t]
\centering
\caption{Ablation study of DG-GSS. Dice scores are reported in \%.}
\label{tab:ablation_dggss}
\scriptsize
\begin{tabular}{lcccccc}
\hline
Variant & Params(M) & FLOPs(G) & PH$^2$ & ISIC2018 & ClinicDB & ColonDB \\
\hline
MKIR baseline (MK-UNet) & 0.316 & 0.314 & 94.39 & 88.74 & 93.48 & 90.01 \\
w/ GroupedSS2D & 0.416 & 0.281 & 94.47 & 88.79 & 93.76 & 90.64 \\
w/ DG-GSS & 0.493 & 0.322 & 94.90 & 88.91 & 94.04 & 91.24 \\
\hline
\end{tabular}
\end{table*}

\subsection{Datasets and Metrics}

We evaluate GAD-MambaUNet on four public binary medical image segmentation datasets covering two representative segmentation tasks. For skin lesion segmentation, we use PH$^2$~\cite{mendoncca2013ph}, which contains 200 dermoscopic images with expert lesion annotations, and ISIC2018~\cite{codella2019skin}, which contains 2,594 dermoscopic images with pixel-level lesion masks. For polyp segmentation, we use CVC-ClinicDB~\cite{bernal2015clinicdb}, which contains 612 colonoscopic polyp images, and CVC-ColonDB~\cite{tajbakhsh2016colondb,vazquez2017benchmark}, which contains 379 colonoscopic images collected from different video sequences. These datasets include diverse imaging conditions, lesion appearances, object scales, boundary ambiguities, and foreground--background contrasts, providing a comprehensive evaluation of the robustness of lightweight segmentation models.

We split each dataset into training, validation, and test subsets with a ratio of 80:10:10. The validation set is used for model selection, and the test set is used for final performance reporting. Dice score is used as the primary segmentation metric, and Intersection over Union (IoU) is also reported when applicable. To evaluate model efficiency, we report the number of trainable parameters and floating-point operations (FLOPs). Since the DINOv3 teacher is used only during training and is discarded during deployment, it does not introduce additional inference computation. Therefore, all efficiency metrics are computed using the student network alone.

\subsection{Implementation Protocol}

All experiments are implemented in PyTorch and conducted under the same training and evaluation settings for fair comparison. Following the input-resolution settings adopted in MK-UNet, images from PH$^2$ and ISIC2018 are resized to $256\times256$, while images from CVC-ClinicDB and CVC-ColonDB are resized to $352\times352$. The validation set is used for checkpoint selection, and the corresponding test performance is reported.

For GAD-MambaUNet, we use AdamW as the optimizer with an initial learning rate of $5\times10^{-4}$, a weight decay of $10^{-4}$, and cosine learning-rate decay to $10^{-6}$. All models are trained for 400 epochs with a batch size of 8, gradient clipping at 0.5, and standard data augmentation. The segmentation objective is the sum of weighted binary cross-entropy and weighted IoU losses. Unless otherwise specified, each configuration is repeated over five independent runs, and the mean test performance is reported.

For training-time semantic supervision, we use frozen DINOv3 ViT-T/16 and ViT-B/16 models as semantic teachers in different experimental settings. The student feature after the first decoder block is aligned to the teacher feature through the projection head described in Section~\ref{sec:method}. DINOv3 distillation starts at epoch 10 and is linearly warmed up for 20 epochs. GAD is activated at epoch 30 and adaptively controls the distillation coefficient until epoch 360. During inference, the teacher branch, projection head, and GAD update are discarded, so the reported parameters and FLOPs are computed using only the student network.

\subsection{Comparison with Reference Methods}

Tab.~\ref{tab:reference_results} compares GAD-MambaUNet with representative medical image segmentation networks and lightweight baselines on four public datasets. Compared with heavy encoder--decoder and Transformer-based models, the proposed models achieve competitive or better accuracy with substantially lower complexity. As shown in Tab.~\ref{tab:reference_results}, TransUNet requires 105.32M parameters and 38.52G FLOPs with an average Dice of 91.70\%, whereas GAD-MambaUNet-B without DINOv3 achieves a higher average Dice of 92.27\% using only 0.493M parameters and 0.322G FLOPs. This indicates that the proposed asymmetric local--global design can improve segmentation accuracy without relying on a heavy backbone.

Among lightweight methods, MK-UNet is a strong reference baseline because it achieves competitive results with compact multi-kernel convolutional blocks. Compared with MK-UNet, which obtains an average Dice of 91.66\% with 0.316M parameters and 0.314G FLOPs, GAD-MambaUNet-B without DINOv3 improves the average Dice to 92.27\% with comparable computational cost. This gain is mainly attributed to introducing DG-GSS blocks only at deep low-resolution stages, thereby enhancing contextual modeling while keeping the inference cost low. Specifically, GAD-MambaUNet-B obtains 94.90\%, 88.91\%, 94.04\%, and 91.24\% Dice on PH$^2$, ISIC2018, CVC-ClinicDB, and CVC-ColonDB, respectively. Increasing the model scale further improves the ClinicDB result, with GAD-MambaUNet-Large achieving 94.70\% on CVC-ClinicDB and 91.37\% on CVC-ColonDB.

Training-time DINOv3 supervision consistently improves the student models without changing their student-side inference complexity. For example, GAD-MambaUNet-Base improves from 94.90\% to 95.1\% on PH$^2$, from 88.91\% to 89.06\% on ISIC2018, from 94.04\% to 94.17\% on CVC-ClinicDB, and from 91.24\% to 91.96\% on CVC-ColonDB. The improvement is more evident on the challenging CVC-ColonDB dataset, where stronger semantic guidance helps distinguish polyp regions from visually similar background tissues. The Medium variant with DINOv3 obtains the best ColonDB Dice of 93.08\%, while the Large variant with DINOv3 achieves the best PH$^2$, ISIC2018, and CVC-ClinicDB results among the compared variants.

Overall, the results show that DG-GSS improves the accuracy--efficiency balance of the lightweight student, and DINOv3-GAD further enhances semantic representation during training without increasing deployment cost. This supports the motivation of combining deep direction--group state-space interaction with training-time foundation-model supervision for lightweight medical image segmentation.

\subsection{Ablation Study of DG-GSS}




To verify the effectiveness of the proposed Direction-Group Graph Selective Scan, we conduct an ablation study by comparing three representative variants, as shown in Tab.~\ref{tab:ablation_dggss}. The MKIR baseline~\cite{rahman2025mkunet} corresponds to the lightweight convolutional design of MK-UNet, which mainly relies on local multi-kernel feature extraction. The GroupedSS2D variant replaces part of the deep local blocks with grouped multi-directional selective scanning, while the complete DG-GSS variant further introduces direction--group graph interaction among scan-direction and channel-group responses.

Compared with the MKIR baseline, introducing GroupedSS2D improves the Dice score from 94.39\% to 94.47\% on PH$^2$, from 88.74\% to 88.79\% on ISIC2018, from 93.48\% to 93.76\% on CVC-ClinicDB, and from 90.01\% to 90.64\% on CVC-ColonDB. This indicates that replacing deep local convolutional blocks with grouped selective scanning can enhance long-range contextual modeling while maintaining low computational cost.

The complete DG-GSS variant further improves over GroupedSS2D on all four datasets, achieving 94.90\%, 88.91\%, 94.04\%, and 91.24\% Dice on PH$^2$, ISIC2018, CVC-ClinicDB, and CVC-ColonDB, respectively. The improvement is especially clear on CVC-ColonDB, where DG-GSS improves the Dice score by 0.60\% over GroupedSS2D and by 1.23\% over the MKIR baseline. This suggests that graph-based interaction among direction--group responses helps capture complementary contextual cues across scan directions and channel subspaces.

In terms of efficiency, DG-GSS keeps the computation lightweight by applying state-space modeling only at low-resolution deep stages. Although the parameter count differs across variants, all models remain within the lightweight range, and DG-GSS achieves the best overall segmentation performance with only 0.493M parameters and 0.322G FLOPs. These results show that the proposed DG-GSS module improves the accuracy--efficiency balance of the student network rather than simply relying on a heavier architecture.

\section{Conclusion}

In this paper, we proposed GAD-MambaUNet, a lightweight medical image segmentation network that combines efficient local modeling, direction--group state-space interaction, and training-time foundation-model supervision. To improve contextual modeling in compact segmentation networks, we introduced Direction-Group Graph Selective Scan (DG-GSS), which treated scan-direction and channel-group responses as graph nodes and enabled structured information exchange before multi-directional fusion. We further incorporated DINOv3-GAD supervision, where a frozen DINOv3 teacher provided semantic guidance during training, and Gradient-Adaptive Distillation dynamically regulated the distillation strength. GAD-MambaUNet achieves a favorable accuracy--efficiency balance compared with representative lightweight and general segmentation methods. Ablation studies further verify the effectiveness of DG-GSS and training-time DINOv3-GAD supervision. In future work, we will explore more flexible teacher--student alignment strategies and extend the proposed framework to more diverse medical segmentation scenarios, such as multi-class and multi-modal segmentation tasks.

\bibliographystyle{model1-num-names}
\bibliography{paper_refs}

\end{document}